**This is the Accepted Manuscript version of an article published by De Gruyter in *Corpus Linguistics and Linguistic Theory* on February 23, 2026.**

The Version of Record is available online at: **https://doi.org/10.1515/cllt-2025-0046**

---

Kaja Dobrovoljc

# Counting trees: a treebank-driven exploration of syntactic variation in speech and writing across languages

University of Ljubljana, Slovenia
Jozef Stefan Institute, Ljubljana, Slovenia

**Abstract:** This paper presents a novel treebank-driven approach to comparing syntactic structures in speech and writing using dependency-parsed corpora. Adopting a fully inductive, bottom-up method, we define syntactic structures as delexicalized dependency (sub)trees and extract them from spoken and written Universal Dependencies (UD) treebanks in two syntactically distinct languages, English and Slovenian. For each corpus, we analyze the size, diversity, and distribution of syntactic inventories, their overlap across modalities, and the structures most characteristic of speech. Results show that, across both languages, spoken corpora contain fewer and less diverse syntactic structures than their written counterparts, with consistent cross-linguistic preferences for certain structural types across modalities. Strikingly, the overlap between spoken and written syntactic inventories is very limited: most structures attested in speech do not occur in writing, pointing to modality-specific preferences in syntactic organization that reflect the distinct demands of real-time interaction and elaborated writing. This contrast is further supported by a keyness analysis of the most frequent speech-specific structures, which highlights patterns associated with interactivity, context-grounding, and economy of expression. We argue that this scalable, language-independent framework offers a useful general method for systematically studying syntactic variation across corpora, laying the groundwork for more comprehensive data-driven theories of grammar in use.

# 1 Introduction

Based on the unitary approach to the study of language, whereby speech and writing are seen as two ends of the same continuum that needs to be described as a whole (Leech 2000; Sinclair and Mauranen 2006), the past four decades have witnessed an unprecedented increase of spoken language research aimed at describing speech-specific linguistic phenomena, which have been ignored or insufficiently addressed by traditional grammatical frameworks (Carter and McCarthy 2017). Although spoken communication exhibits particularities on various levels of linguistic description due to the specific circumstances of its production (Biber 1988; Chafe and Tannen 1987; Halliday 1989), it is most notably characterized by its distinct syntactic behavior involving features such as disfluencies, ellipsis, and parenthetical constructions, to name just a few (Biber et al. 1999).

The increasing interest in spoken language phenomena has partially been motivated by the positivistic, data-driven methodological developments in linguistics in general, with the rise of spoken corpora, large digitized collections of transcribed speech, allowing linguists to discover aspects of language that were not previously accessible to empirical observation (Adolphs and Carter 2013; Stubbs and Halbe 2012). Their comparison with written corpora has enabled a systematic exploration of similarities and differences between speech and writing, particularly through bottom-up, corpus-driven approaches, which allow linguistic patterns to emerge inductively from data rather than relying on predefined assumptions (Biber 2015; Tognini-Bonelli 2001). While these methods have uncovered valuable modality-specific contrasts, they have largely remained focused on surface-level features, such as words (Leech and Rayson 2014), lexical bundles (Biber 2009; Biber et al. 2004), or part-of-speech tags (Hardie 2007) and thus offer limited access to the full structural complexity of speech.

In parallel, the past few decades have also seen the emergence of linguistically annotated corpora, which enrich raw text with additional layers of linguistic information and thus enable far more complex analyses of corpus data than investigations of lexical patterns alone (Gries and Berez 2017; Ide and Pustejovsky 2017; Kübler and Zinsmeister 2014). Of particular relevance to syntactic research are morphosyntactically parsed corpora, or treebanks, in which surface forms are annotated for both morphological and syntactic features. Although such corpora have traditionally been developed on written texts, they are now increasingly available for speech, with a growing number of spoken treebanks providing syntactic analyses of transcribed spontaneous interaction across a wide range of languages (e.g., Godfrey et al. 1992;

Hajič et al. 2008; Hinrichs et al. 2000; Kåsen et al. 2022; Lacheret-Dujour et al. 2019; MacWhinney 2014; Øvrelid et al. 2018; Schuurman et al. 2003, to list a few).

However, despite the growing availability of syntactically annotated spoken corpora, their full methodological potential for advancing our understanding of spoken grammar – particularly in comparison to writing – remains largely underexplored. Most comparative studies using spoken treebanks have focused on specific syntactic constructions (Hüll and Dobrovoljc 2025; Kyle and Eguchi 2023; Levshina et al. 2023; Pietrandrea and Delsart 2019; Roland et al. 2007; Van Eynde 2009), the distribution of part-of-speech or syntactic labels (Hinrichs and Kübler 2005; Poiret and Liu 2020), or on summary metrics such as syntactic complexity (Wang and Liu 2017). This research-goal-specific approach to investigating parsed data is also characteristic of treebank-based research more generally, where studies typically examine predefined subsets of configurations motivated by particular theoretical or applied interests. For instance, studies have focused on local subtrees and syntactic triples (Futrell et al. 2015; Goldberg and Orwant 2013; Sidorov 2019), combinations of specific part-of-speech tags and syntactic labels (Díez-Bedmar and Pérez-Paredes 2020; Gerdes et al. 2021; Kyle and Crossley 2017; Levshina 2019; Wong et al. 2017), parsed representations of full sentence structures (Liu et al. 2022) and other predefined production units (Jiang et al. 2019). While these targeted approaches have yielded valuable insights into specific phenomena, they make only partial use of the structural richness encoded in syntactic annotations and stop short of systematically extracting and comparing the *full* range of syntactic structures attested in the data. As a result, we still lack a comprehensive understanding of how syntactic patterns vary across modalities, particularly when approached through fully inductive, hypothesis-free extraction of all attested structural configurations**.**

To address this gap, the present study proposes an exhaustive bottom-up, corpus-driven approach to identifying and analyzing syntactic similarities and differences between speech and writing by leveraging the rich syntactic information encoded in dependency-parsed corpora. Specifically, we take syntactic structures—defined as the complete set of dependency trees and subtrees extracted from each corpus—as our core unit of analysis, and systematically compare both their types (unique structures) and tokens (frequencies) across modalities to answer the following research questions:

**RQ1.** How does the inventory of syntactic structures in speech vary in comparison to writing?
**RQ2**. To what extent do spoken and written language share the same syntactic structures?
**RQ3**. Which syntactic structures are particularly characteristic of speech?

In doing so, we develop a systematic methodological framework for corpus-driven syntactic analysis, which combines exhaustive syntactic structure extraction with distributional comparison techniques. Crucially, the proposed framework is language-independent and applicable to any pair of syntactically parsed corpora. We demonstrate this potential through a cross-linguistic case study of English and Slovenian—two typologically distinct languages with parallel spoken and written treebank data annotated under the same annotation framework.

The remainder of this paper is structured as follows. Section 2 introduces the corpora used in this study and outlines the syntactic annotation framework on which they are based. Section 3 defines our core units of analysis as full dependency trees and subtrees, explains the procedure for extracting these structures from parsed data, and presents the methods used to compare their distributions across corpora. Section 4 reports our findings on syntactic variation across spoken and written corpora, highlighting cross-modal differences in the diversity, composition and overlap of syntactic inventories. Section 5 discusses the methodological and linguistic implications of these findings, concluding with a reflection on the broader applicability of our framework for corpus-driven studies of syntactic variation in general.

# 2 Corpora

To illustrate the language-independent nature of our approach and control for language-specific factors in the interpretation of results, this study draws on spoken and written corpora from two syntactically distinct languages: English and Slovenian. The corpora are described in detail in Sections 2.1 and 2.2, while Section 2.3 introduces their shared annotation framework – Universal Dependencies (UD) – which ensures comparability across both modalities and languages. Section 2.4 presents structurally normalized corpus versions designed to minimize modality-specific artifacts, and Section 2.5 outlines key typological differences between the two languages that provide context for interpreting cross-linguistic patterns.

## 2.1 GUM treebank of written and spoken English

The Georgetown University Multilayer (GUM) corpus (Zeldes 2017) is a freely available English corpus of richly annotated texts from various genres, providing a diverse representation of contemporary English. GUM features multiple layers of linguistic annotation, including morphosyntactic annotations following the Universal Dependencies (UD) scheme. This layer of the corpus, also known as the English GUM Treebank, is included in the regular semi-annual releases of the UD multilingual treebank collection. For this study, we used the GUM version included in UD v2.15 (Zeman et al. 2024), which contains 12,146 manually parsed sentences and a total of 211,920 words.

To support the cross-modal comparison introduced in the introduction, we divided the GUM Treebank data into spoken and written subsets using the available genre classification: the GUM-spoken subset includes data from interviews (22% of the words in the corpus), vlogs (21%), speeches (21%), conversations (20%), courtroom transcripts (9%), and podcasts (7%), while the GUM-written subset comprises biographies (14%), fiction (13%), academic texts (13%), textbooks (13%), how-to guides (13%), news articles (12%), travel guides (12%), letters (5%), and essays (4%). Detailed statistics for each subset, including token counts without punctuation and disfluencies (see Section 2.4), are provided in Table 1.

**Table 1:** Basic statistics for the spoken and written subsets of the English GUM Treebank.

| Treebank | Documents | Sentences | Words | Words (no punct) | Words (no disfluency) |
|---|---|---|---|---|---|
| GUM-written | 143 | 6,493 | 130,990 | 113,354 | 113,199 |
| GUM-spoken | 74 | 5,653 | 80,930 | 69,611 | 67,031 |
| Total English | 217 | 12,146 | 211,920 | 182,965 | 180,230 |

## 2.2 SSJ and SST treebanks of written and spoken Slovenian

In contrast to the multi-genre GUM Treebank, spoken and written data for Slovenian have been developed in two separate projects and are released as distinct treebanks. The SST Treebank of Spoken Slovenian (Dobrovoljc and Nivre 2016; Dobrovoljc 2025c) is a manually syntactically annotated sample of the *GOS* reference corpus (Verdonik et al. 2024), containing transcribed monologic, dialogic, and multi-party speech from everyday situations, robustly classified as public informative or educational discourse (42%, e.g. news broadcast, lectures, school lessons), public entertainment (12%, e.g. talk shows and interviews), non-public conversations (10%, e.g. workplace, services, sales), and private conversations between friends or family (38%). It is balanced for speaker demographics (sex, age, region, education), communication channels (TV, radio, telephone, in-person), and communicative settings (e.g., broadcasts, lectures, meetings, consultations, informal conversations).

Its written counterpart, the SSJ reference Slovenian treebank (Dobrovoljc et al. 2017; Dobrovoljc et al. 2022), represents a balanced and representative manually annotated sample of the Gigafida reference corpus of standard written Slovenian (Krek et al. 2020). As such, it encompasses a wide variety of genres from diverse domains and publishers, which are broadly classified as newspaper (39%) and magazine (22%) articles, instructional texts (15%), such as textbooks, guides and manuals, Wikipedia articles (12%), fiction (8%), online news (3.5%) and other web content (0.5%). For both treebanks, we used the version included in UD release v2.15 (Zeman et al. 2024). Detailed statistics for each treebank, including token counts with and without punctuation (see Section 2.4), are provided in Table 2.

**Table 2:** Basic statistics for the spoken and written Slovenian treebanks.

| Treebank | Documents | Sentences | Words | Words (no punct) | Words (no disfluency) |
|---|---|---|---|---|---|
| SSJ (written) | 618 | 13,435 | 267,097 | 227,621 | 227,421 |
| SST (spoken) | 344 | 6,108 | 98,396 | 76,341 | 68,281 |
| Total Slovenian | 962 | 19,543 | 365,493 | 303,962 | 295,702 |

## 2.3 Universal Dependencies annotation scheme

Universal Dependencies (UD; de Marneffe et al. 2021) is an international initiative launched in 2013 to develop cross-linguistically consistent grammatical annotations for many languages. Building upon earlier standardization efforts (de Marneffe et al. 2014; Petrov et al. 2012; Zeman 2008), UD provides a universal set of categories and annotation guidelines, ensuring consistent representation of similar syntactic constructions across languages while allowing for language-specific extensions when necessary. The framework includes 17 universal part-of-speech (POS) tags (e.g., `NOUN` for nouns), 24 morphological features formalized as attribute-value pairs (e.g., `Number=Sing` for singular number), and 37 dependency relations (e.g., `nsubj` for nominal subject). For a full list of UD POS tags and syntactic relations, please refer to Appendix A (Tables A1 and A2).

At the syntactic level, UD follows the principles of dependency grammar (Tesnière 1959; Mel′čuk 1988), where each word in a sentence is linked to a governing head via a labeled syntactic relation. These head-dependent links form a dependency tree, with a single root (typically the main predicate) and all other words connected through binary relations such as subject, object, or modifier. Unlike phrase-structure grammars that group words into nested phrases, dependency grammars thus establish direct relationships between words, making them especially well suited for representing syntactic structure across languages with both fixed and flexible word orders, where heads and dependents may be far apart or appear in variable positions.

An example of a dependency tree is given in Figure 1, which shows the UD analysis of the GUM sentence *She stayed while I lit the fire*. In the tree, each word is connected to its syntactic head via a labeled arc, forming a hierarchical structure rooted in the main verb *stayed*. The subject *She* is linked to *stayed* with the relation `nsubj` (nominal subject), while the subordinate clause *while I lit the fire* is attached as an `advcl` (adverbial clause modifier). Within that clause, *lit* is the head verb, governing *I* as its `nsubj`, *fire* as its `obj` (object), and *the* as a `det` (determiner) modifying *fire*. The word *while* is attached to *lit* with the `mark` relation, indicating its role as a clause-introducing marker.

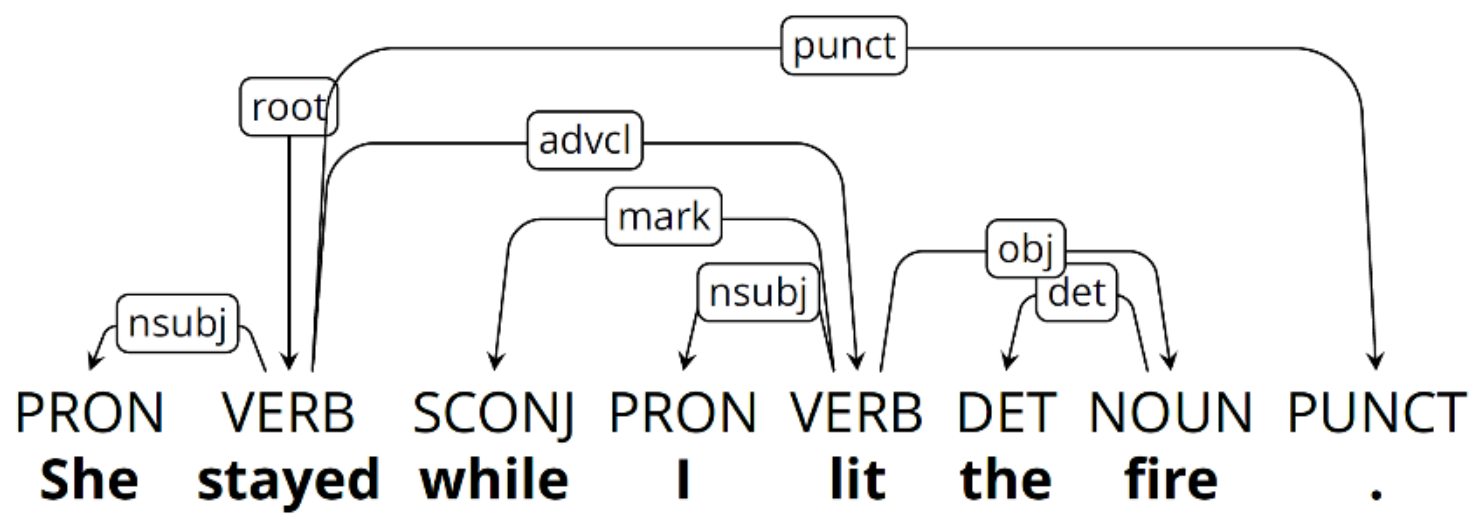

**Figure 1:** Dependency tree of the GUM written sentence 'She stayed while I lit the fire.', annotated according to the Universal Dependencies scheme.

Although primarily developed for written corpora, UD also provides syntactic relations to account for speech-specific phenomena, such as self-repairs (`reparandum`), vocatives (`vocative`), and discourse markers (`discourse`). As illustrated in the spoken example in Figure 2, this enables a comprehensive, single-layer analysis of spoken transcriptions, including features such as disfluencies and interactional markers that were often excluded in earlier syntactic annotation frameworks. At the same time, these phenomena are encoded in a way that allows for their easy exclusion when needed, offering flexibility for different types of linguistic analysis—a point we illustrate in Section 2.4 and revisit in the discussion (Section 5).

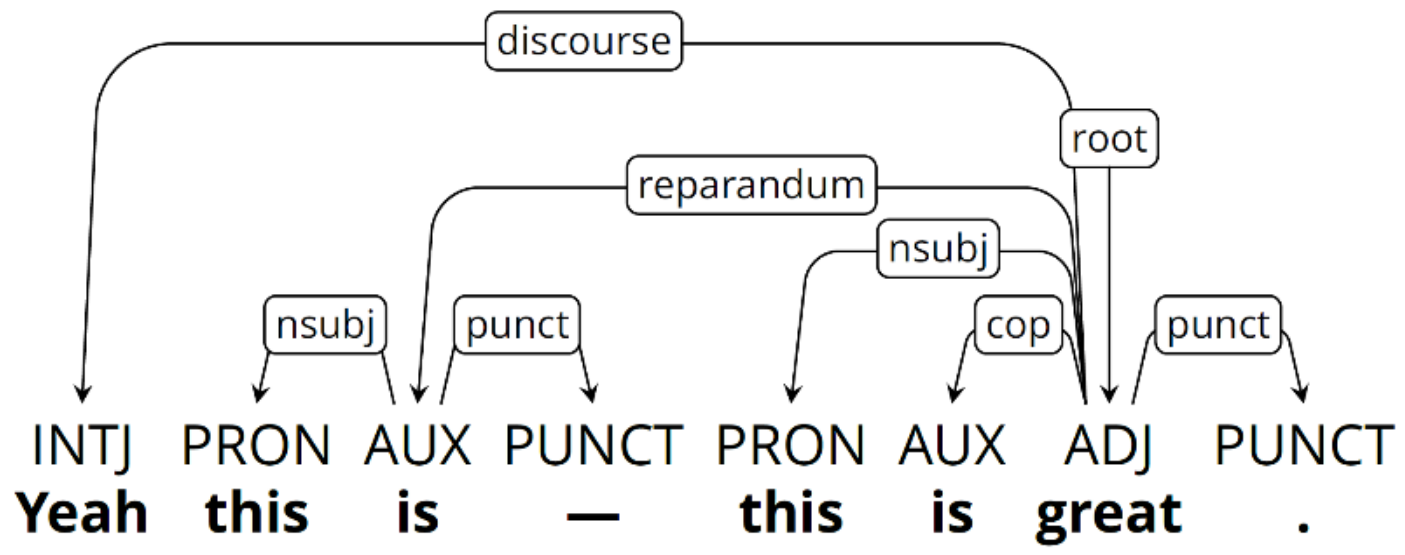

**Figure 2:** Dependency tree of the GUM spoken utterance 'Yeah this is – this is great.', annotated according to the Universal Dependencies scheme.

## 2.4 Structurally normalized corpus versions

Although we acknowledge that speech and writing are not monolithic modalities—many genres occupy intermediate positions between them, including *conceptually spoken* writing such as dialogue in fiction and *conceptually written* speech such as pre-prepared speeches (Koch and Oesterreicher 1985; Leuckert and Buschfeld 2021)—we nevertheless operationalize this distinction in practical terms. To minimize the influence of modality-specific idiosyncrasies and ensure a structurally focused and directly comparable analysis across both modalities, we created two normalized versions of each corpus: a punctuation-free version and a disfluency-free version.

The punctuation-free version serves as the primary basis for all analyses. In spoken corpora, punctuation is a transcriptional artifact rather than part of the original utterance, which would introduce an asymmetry in comparison with written data. An illustrative example is provided in Figure 3, showing the parsed sentence from Figure 2 after punctuation removal.

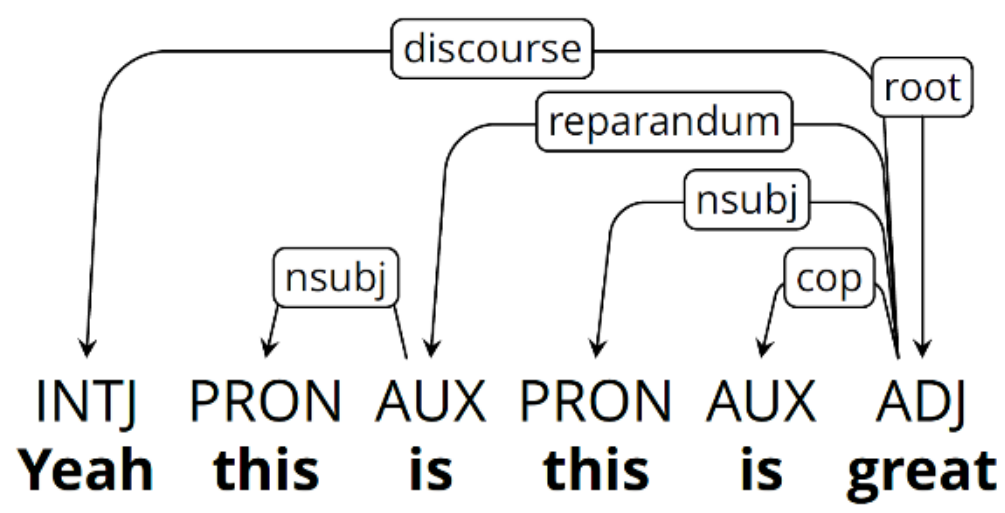

**Figure 3:** Dependency tree of the example in Figure 2 after removal of branches labeled as *punct* (punctuation).

To further ensure the robustness of our findings—that is, to verify that the syntactic observations presented in Section 4 are not merely driven by performance-related features such as self-repairs, false starts, filled pauses and other types of disfluencies—we additionally created a disfluency-free version of all corpora, in which we removed not only punctuation (`punct`) but also all branches rooted in tokens labeled as `reparandum` and `discourse`, as illustrated in Figure 4. By minimizing the influence of elements typical of spontaneous speech, this version enables a sharper focus on core syntactic constructions.

Both versions—the punctuation-free and the disfluency-free corpora—were used in the study, with the main text and appendix presenting complementary results depending on the type of analysis. Updated token counts for both versions are reported in Tables 1 and 2, alongside the original corpus statistics.

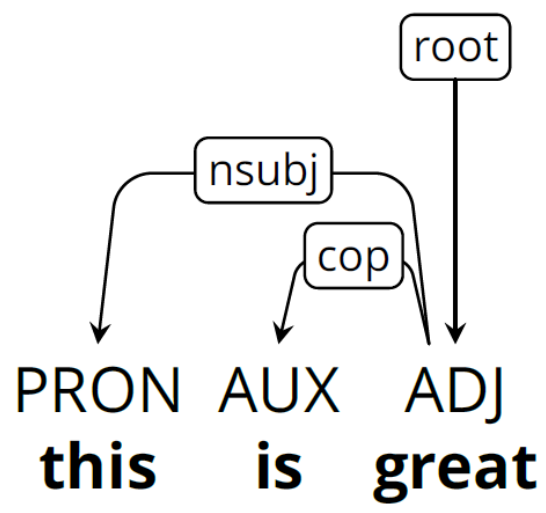

**Figure 4:** Dependency tree of the example in Figure 2 after removal of branches labeled as *punct* (punctuation), *discourse* (discourse elements) or *reparandum* (self-repairs).

## 2.5 Typological background: English and Slovenian

Finally, to provide context for interpreting cross-linguistic patterns, we briefly outline key typological differences between English and Slovenian.

Slovenian is a South Slavic language with rich inflectional morphology, including a six-case system, complex verbal conjugation, and three grammatical genders. It exhibits relatively flexible word order, where constituent ordering is primarily governed by information structure and pragmatic considerations rather than grammatical function. Additionally, Slovenian is a pro-drop language, allowing subject pronouns to be omitted when recoverable from context—which introduces an additional source of structural variation, as many trees with verbal roots can appear with or without an overt subject. In contrast, English has minimal inflectional morphology and relies primarily on word order to mark grammatical relations, with more constrained possibilities for constituent reordering and obligatory subject expression.

These typological differences mean that the two languages may achieve similar communicative functions through different structural means, which could surface as distinct patterns in extracted dependency inventories. For instance, Slovenian's word order flexibility might be expected to generate greater structural diversity than English in both spoken and written language. However, the primary value of including both languages is to demonstrate the language-independence and broad applicability of the proposed method.

# 3 Syntactic structure extraction and comparison

In the next stage of our analysis, we extracted all syntactic structures observed in each corpus, along with their frequencies, in order to compare their distribution across modalities. This involved first establishing a working definition of 'syntactic structure', treating each dependency (sub)tree as a quantifiable unit of analysis (Section 3.1). We then used the STARK tool to automatically extract these structures from each corpus using a uniform set of parameters (Section 3.2). Finally, we applied a set of quantitative methods to compare the extracted structures across corpora (Section 3.3).

## 3.1 Defining syntactic structures as dependency (sub)trees

As already outlined in the introduction, our goal was to move beyond prior work that primarily focused on the distribution of isolated syntactic categories (e.g. POS tags or dependency labels) or predefined constructions (e.g. predicate-object bigrams), and instead adopt a fully corpus-driven approach that establishes and examines the full set of syntactic structures attested in a corpus. Following this approach, we define syntactic structures as all (labeled) dependency trees and subtrees attested in a treebank. More specifically, we consider each word in the corpus as the head (root) of its own syntactic structure (dependency tree or subtree),[1] comprising the word itself along with all its direct and indirect dependents. This bottom-up, treebank-driven approach thus captures structures of any size- from single-word trees to full sentence-level trees —without relying on externally imposed criteria for what constitues a syntactic unit of interest.

To illustrate this process more concretely, consider the sentence shown in Figure 1 (presented without punctuation). As summarized in Table 3, the set of structures identified in this sentence yields seven distinct trees, each rooted at a different node in the original dependency graph—ranging from the root down to individual tokens (leaves). Each of these is treated as a separate *syntactic structure*: this includes the full sentence tree (rooted at the main verb *stayed*), intermediate trees such as the adverbial clause (rooted at *lit*) and the direct object (rooted at *fire*), and single-word trees for nodes with no dependents (*She, while, I, the*). By systematically traversing each sentence in this way, we obtain the full set of syntactic configurations attested in the corpus, which we treat as a finite empirical inventory for analysis.

[1] In the remainder of this paper, we use the term *dependency tree* to refer to both full trees (i.e., those spanning from the *root* node—typically the predicate of the main clause—and encompassing the entire sentence) and subtrees (those rooted in any other word in the sentence).

To support abstract syntactic analysis and facilitate generalization, we represent each tree in a delexicalized form. For the primary analysis, we follow the prevailing approach in related work (see discussion by Futrell et al. 2015: 95) and use a full morphosyntactic representation in which nodes are defined by their part-of-speech tags (e.g., ADJ, NOUN, PRON, VERB), while branches encode dependency relations (e.g., nsubj, obj, amod). Table 3 shows the full set of extracted structures from the example sentence, displayed as such delexicalized trees.

**Table 3:** Syntactic structures (trees) in the sentence *She stayed while I lit the fire* (Figure 1).

| Head word | Dependency (sub)tree | Lexicalized |
|---|---|---|
| **She** | ↓ PRON | *She* |
| **stayed** | 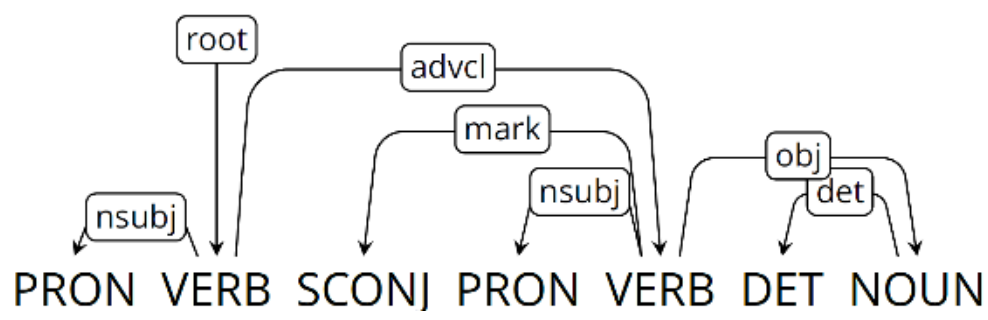 | *She stayed while I lit the fire* |
| **while** | ↓ SCONJ | *while* |
| **I** | ↓ PRON | *I* |
| **lit** | 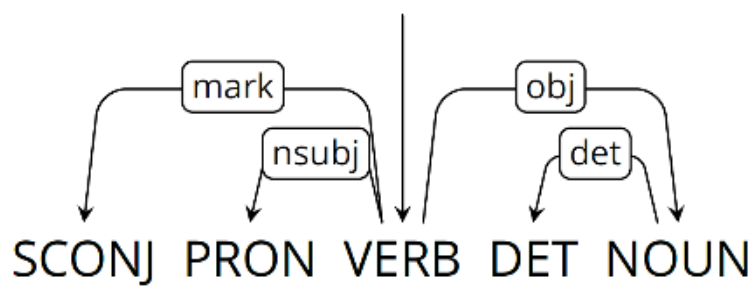 | *while I lit the fire* |
| **the** | ↓ DET | *the* |
| **fire** | 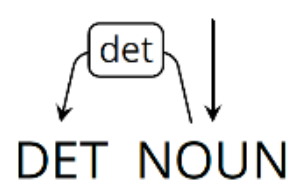 | *the fire* |

In addition to the full morphosyntactic representation illustrated in Table 3, we also extract trees using a dependency-only representation, in which nodes are defined solely by dependency relations, collapsing all part-of-speech distinctions. This complementary approach allows us to assess the robustness of our findings across different levels of abstraction. To illustrate this distinction, hypothetical structures like *she stayed* (PRON-VERB), *Mary stayed* (PROPN-VERB), and *women stayed* (NOUN-VERB) would be treated as three distinct subject-predicate structures in the full morphosyntactic form, but as the same structure in the dependency-only form (a single structure with the subject-predicate relation). Examples in the continuation of this section and all primary results are presented using the full morphosyntactic representation, with parallel findings for the dependency-only representation provided in Appendix C. Where relevant, we comment on noteworthy similarities or differences between the two representations throughout Section 4.

## 3.2 Extracting syntactic structures with STARK

To extract the delexicalized dependency structures described in the previous section, we used STARK (Krsnik et al. 2025),[2] a tool for statistical treebank analysis designed to support both targeted and exploratory queries. STARK can retrieve specific structures based on predefined criteria (top-down tree querying) or extract all structures in a treebank without prior assumptions (bottom-up tree extraction), aligning well with the corpus-driven methodology adopted in this study.

For our purposes, we configured STARK to extract labeled, delexicalized trees of any size, treating word order as a structurally distinctive feature—so that, for instance, the premodifying *beautiful day* (ADJ < NOUN) and the postmodifying *court martial* (NOUN > ADJ) are treated as two distinct structures. To ensure full cross-linguistic comparability, we collapsed fine-grained label subtypes under their core label types (e.g., `cc:preconj` is treated simply as `cc`). The inclusion of such subtypes could, however, serve as a valuable extension in future work, especially when working with corpora where genre-specific or idiosyncratic phenomena are annotated systematically and comparably – as, for example, speech-related extensions proposed by Schneider and Zeldes (2021), Kahane et el. (2021) and Dobrovoljc (2022).[3]

---

[2] https://github.com/clarinsi/STARK

[3] The full STARK configuration file is available in the accompanying data release (Dobrovoljc 2025a, 2025b). Key settings include: `node_type = upos` (part-of-speech tags as nodes), `labelled = yes` (include dependency labels), `label_subtypes = no` (ignore subtypes), `fixed = yes` (preserve word order). For the dependency-only representation, the configuration is identical except for `node_type = None` (nodes without part-of-speech labels).

The tool was applied to each of the four corpora described in Section 2, using their punctuation-free versions introduced in Section 2.4. For each corpus, the output consists of a tabular file listing all extracted structures (one tree per row), along with their frequency and additional metadata. Each extracted tree is represented as a one-line string inspired by the dep_search query language (Luotolahti et al. 2017)— a compact, human-readable notation well suited for both automated analysis and manual inspection. In this format:

- **Dependencies** are expressed using < and >, which mimic arrows in dependency graphs:
  - A < B means token A is governed by token B (e.g., *rainy < morning*)
  - A > B means token A governs token B (e.g. *read > newspapers*)
- **Dependency labels** follow the operator:
  - A <amod B means token A is an adjectival modifier of B (e.g., *rainy <amod morning)*
  - A >obj B means B is the direct object of A (e.g., *read >obj newspapers*)
- **Structural priority** is marked with parentheses:
  - A > B > C means A governs both B and C in parallel (e.g., *people < read > newspapers*)
  - A > (B > C) means A governs B which, in turn governs C (e.g., *read > (interesting < newspapers)*)

For illustration, Table 4 shows the most frequent types of NOUN-headed trees extracted from the GUM-written corpus using STARK, i.e. the list of most frequent types of nominal phrase structures found in English written texts. This notation is used throughout the paper to describe and compare syntactic structures under investigation.

**Table 4:** Top-5 most frequent types of NOUN-headed trees in the written subset of the English GUM treebank. The first structure indicating embedded dependents within a nominal phrase (i.e. involving bracketing) appears in 20th position as DET <det NOUN >nmod (ADP <case NOUN) (e.g., *a shortage of manpower*).

| Tree | Example | Frequency |
|---|---|---|
| NOUN | *car* | 3638 |
| DET <det NOUN | *the car* | 1507 |
| ADP <case NOUN | *by car* | 1419 |
| ADP <case DET <det NOUN | *with a car* | 1342 |
| ADJ <amod NOUN | *fast car* | 636 |

With this procedure, we extracted the complete set of syntactic structures attested in each of the four corpora, which serves as the empirical basis for all analyses presented in the remainder of the paper. For transparency and reusability, the full lists of extracted structures are available in an open-access repository (Dobrovoljc 2025a, 2025b).

## 3.3 Methods for comparing syntactic structures

The extracted lists of syntactic structures from the spoken and written corpora, as described in Section 3.2, form the basis for a bottom-up comparison of syntactic usage across modalities. To address the research questions outlined in the introduction, we applied standard corpus-linguistic measures—commonly used in lexical analysis—but here applied to syntactic structures as the linguistic items of interest. Specifically:

- **Comparison of syntactic inventory (RQ1):** To compare the size and variability of syntactic inventories across speech and writing, we report the total number of unique structures in each corpus, compute the type-token ratio, and analyze distributions by structural type (i.e., head part-of-speech).
- **Comparison of syntactic overlap (RQ2):** To assess the extent of structural similarity between spoken and written language, we calculate the number of shared and unique structures.
- **Identification of speech-specific structures (RQ3):** To identify structures that are particularly characteristic of speech, we use %DIFF keyness measure (Gabrielatos and Marchi 2012), a commonly used effect-size metric in corpus linguistics to measure the proportion of the diference between the normalized frequencies of a linguistic phenomenon in two corpora.[4]

Further methodological details and all results are presented in Section 4.

[4] The %DIFF metric is calculated as $((NF_SF - NF_RF) / NF_RF) \times 100$, where NF_SF and NF_RF are normalized frequencies in the spoken and written corpora, respectively. When an item is absent from the written corpus (NF_RF = 0), a proxy value of 1×10−15 was used to avoid division by zero, resulting in a very large positive %DIFF value.

# 4 Results

This section presents the results of our corpus-driven analysis of syntactic structures in spoken and written language, based on the corpora introduced in Section 2 and the extraction and comparison procedure described in Section 3. We address the three research questions outlined in Section 1, beginning with a comparison of the diversity and variation of syntactic inventories across modalities (Section 4.1), followed by an analysis of structural overlap (Section 4.2) and an identification of speech-specific structures (Section 4.3). Unless otherwise noted, we present results based on the full corpora versions excluding punctuation (Section 2.4), using the full morphosyntactic representation of dependency trees (Section 3.1). The corresponding results for the disfluency-free version—further excluding *discourse* and *reparandum* structures—are provided in the Appendix B, while results for the dependency-only representation are provided in Appendix C. Notably, with the exception of top-ranked speech-specific structures (Section 4.3), the findings across all versions are very similar, confirming the robustness of our results regardless of the scope of spoken language syntax or the level of structural abstraction considered**.**

## 4.1 Comparison of syntactic inventory size, diversity and composition

Using the procedure described in Section 3, we extracted the complete inventory of syntactic structures from each corpus. Table 5 provides an overview of the extraction results, showing the total number of structure tokens (instances), types (unique structures), and hapax legomena (structures occurring only once) across all four corpora.

**Table 5:** Overview of extracted syntactic structure inventories in English and Slovenian spoken and written corpora (punctuation-free versions): tokens, types, and hapax legomena.

| | **English** | | **Slovenian** | |
|---|---|---|---|---|
| | **Spoken** | **Written** | **Spoken** | **Written** |
| **Structure tokens** | 69,611 | 113,354 | 76,341 | 227,621 |
| **Structure types** | 13,429 | 21,759 | 15,284 | 43,143 |
| **Hapax legomena** | 12,234 | 19,899 | 14,236 | 40,248 |

As Table 5 shows, all corpora exhibit large syntactic inventories, with written corpora consistently containing more structure types than their spoken counterparts. However, this pattern largely reflects differences in corpus size: spoken corpora are considerably smaller than written ones in terms of total structure tokens, and the Slovenian written corpus (SSJ) is substantially larger than its English counterpart (GUM-written). Nevertheless, all four corpora display a similarly skewed distribution: over 90% of all types occur only once (hapax legomena), confirming the combinatorial potential of syntax, where even a limited set of grammatical rules can give rise to a vast number of structural configurations through branching, word order, and morphosyntactic variation.

#### **4.1.1** *Comparison of the syntactic inventory diversity*

To enable meaningful comparison of syntactic diversity across corpora of different sizes, we therefore computed the Type-Token Ratio (TTR), a common measure of linguistic diversity. In our case, types refer to distinct syntactic structures, while tokens correspond to their occurrences in the corpus. Since each word in the corpus serves as the head of exactly one structure, the total number of tokens reflects the actual corpus size. However, because TTR increases with corpus size—particularly in syntax, where longer corpora yield more structural realizations—we used Segmented TTR (STTR), which averages TTR over fixed-size segments (1,000 tokens) rather than full corpora. This approach enables size-independent comparison: higher STTR values indicate greater syntactic diversity, while lower values reflect more repetition. For example, an STTR of 0.28 over 1,000 tokens corresponds to roughly 280 distinct structures, meaning that each structure type occurs on average about 3-4 times per segment.

As shown in Figure 5, STTR values are consistently lower in speech than in writing for both languages ($p < 0.001$), indicating that spoken language tends to rely more heavily on a smaller set of frequently reused syntactic structures, while writing draws from a broader and more varied inventory. This pattern likely reflects pressures of real-time planning and information structuring, with prosodic cues such as intonation, stress, and rhythm providing an important mechanism for conveying meaning and maintaining coherence despite reduced structural variation. At the same time, speech makes extensive use of recurrent, modality-specific constructions–such as formulaic greetings, questions, and replies–that serve as conventionalized templates for interaction.

Importantly, the observed lower syntactic diversity of speech relative to writing holds regardless of analytical scope: it persists when we consider only core syntactic building blocks, excluding disfluencies (Table B1 and Figure B2 in Appendix B), and when we collapse part-of-speech distinctions to focus on purely relational syntactic configurations (Table C1 and Figure C2 in Appendix C). For comparison, lexical diversity (computed as STTR over lemmas) shows similar modality contrasts, with spoken corpora exhibiting lower diversity than written corpora in both languages (English spoken: 0.33, written: 0.42; Slovenian spoken: 0.36, written: 0.54).[5]

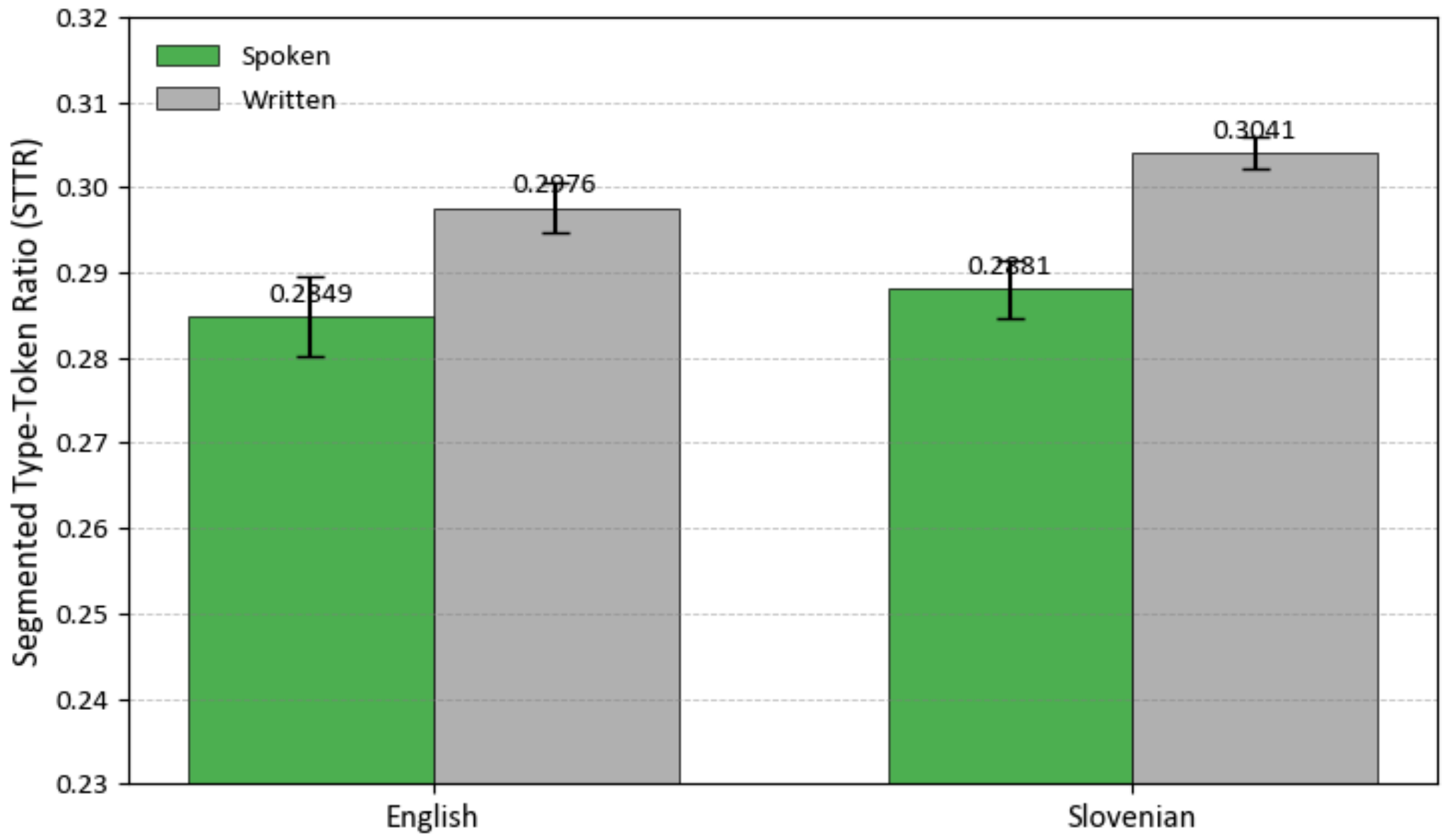

**Figure 5:** Segmented Type-Token Ratio (STTR) in spoken and written treebanks for Slovenian and English. Bars show average TTR values across 1,000-token segments; error bars indicate 95% confidence intervals.

An important caveat, however, is that all the above results pertain to the specific corpora analyzed and do not account for within-corpus and across-language genre composition differences. The potential effect of genre composition is also evident in the dependency-only results (Figure C2), where Slovenian spoken diversity drops substantially more than English. This pattern likely reflects corpus composition differences, as the Slovenian data contain a much larger proportion of spontaneous, unprepared, and interactive speech (about 50 % conversations, compared to 20 % in English, see Sections 2.1 and 2.2), which is typically characterized by more context-dependent vocabulary, including frequent use of pronouns and other deictic expressions (Biber et al. 1999).

Similarly, genre-controlled studies would be needed to assess word order flexibility effects (Section 2.5): while written Slovenian shows consistently higher diversity than English (confirming the typological prediction), spoken results vary by tree representation. In the dependency-only representation (Figure C2), Slovenian spoken diversity drops below English, suggesting that speakers rely on more canonical, repetitive patterns rather than exploiting the full configurational flexibility of the language – an effect likely amplified by the higher proportion of spontaneous, interactive discourse in the Slovenian corpus.

[5] The higher Slovenian figures for lexical diversity likely reflect its productive derivational morphology, which yields distinct lemmas where English uses identical forms (e.g., phrasal verbs like *go (out), go (in)* remain under a single lemma *go*, and noun/verb pairs like *drive* share one form).

#### **4.1.2** *Comparison of the syntactic inventory composition*

To complete our comparison of syntactic inventories, we also examined their internal composition—specifically, how structures are distributed with respect to the part of speech (POS) of their syntactic head. Figure 6 (and Figure B3 for disfluency-free version) visualizes these patterns using a two-dimensional plot, which shows the relative difference in the frequency of each POS in spoken vs. written data, separately for English (x-axis) and Slovenian (y-axis). Each point represents a POS tag; values above zero indicate a higher proportion in speech, while values below zero indicate a higher proportion in writing. Points in the upper-right quadrant thus correspond to POS heads that are more frequent in speech in both languages, while those in the lower-left quadrant indicate POS heads that are more frequent in writing across both languages.

The plot reveals clear modality-driven preferences in both languages, aligning with tendencies described in previous work (e.g., Chafe and Danielwicz 1987; Biber 1988; Halliday 1989; Biber et al. 1999; Leech and Rayson 2014), but shown here systematically and cross-linguistically. Spoken corpora contain a higher proportion of structures headed by pronouns, interjections, particles, and adverbs—categories associated with interactivity, deixis, and discourse management—as well as verbs, reflecting a more dynamic, narrative style. In contrast, written corpora are strongly characterized by noun- and proper noun-headed structures, reflecting the more information-dense and content-oriented nature of writing. They also exhibit more frequent use of prepositions, adjectives, and numerals, suggesting that written language tends to favor more elaborated and hierarchically complex phrases, with greater use of modifiers and embedded elements.

At the same time, the distribution plot in Figure 6 also reveals some interesting language-specific traits. For instance, determiner structures (DET-headed trees) are more common in English writing but more frequent in Slovenian speech. This reflects how definiteness is expressed: in English, it is marked by articles (e.g., *the*, *a*), while in Slovenian, it is mainly encoded morphologically, leaving demonstrative and possessive pronouns (e.g., *ta* ‘this’, *moj* ‘my') as the only determiners, which are naturally more frequent in speech. Conversely, auxiliary structures are notably more frequent in English speech, likely reflecting their frequent use in question formation (e.g., *do you know*) or progressive and perfect aspect (*I'm going, I've been*).

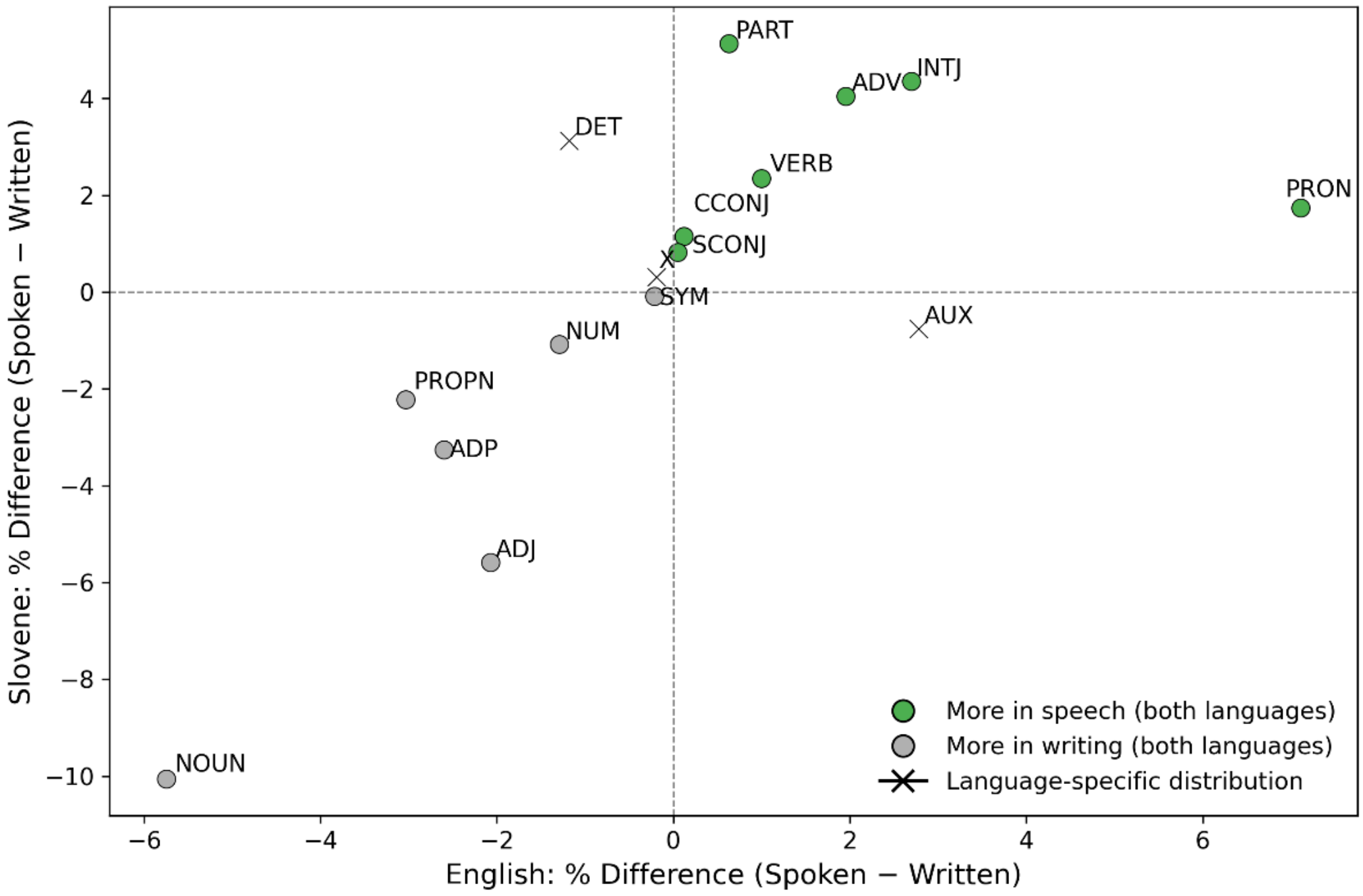

**Figure 6:** Cross-linguistic comparison of structure distributions in spoken and written corpora according to structural type (i.e., the POS tag of the head of the dependency tree). Positive values indicate higher frequency in speech; negative values indicate higher frequency in writing.

## 4.2 Comparison of the syntactic inventory overlap

The preceding section outlined the size, diversity, and internal composition of syntactic inventories in spoken and written corpora, revealing clear modality-specific tendencies. We now turn to a more direct comparison of inventory overlap, which highlights the extent to which spoken and written language share—or diverge in—their structural repertoire.

As shown in Figure 7, the sets of distinct syntactic structures observed in speech and writing are only partially overlapping. In both Slovenian and English, only a small proportion of structures found in the spoken corpora also appear in their written counterparts (11.2% in English; 9.1% in Slovenian), while the majority of structures attested in speech are entirely absent from writing—that is, they are unique to spoken language.

Part of this limited overlap is expected, given the high proportion of hapax legomena in each corpus (see Table 5), which are unlikely to recur across modalities by chance alone.

However, even when we restrict the comparison to more frequent and thus more stable subsets—such as structures occurring at least twice, or the top 200 most frequent structures per corpus—the degree of overlap remains relatively low. For example, only 145 of the top 200 (72.5%) spoken structures in English also occur among the top 200 in writing, and the corresponding number for Slovenian is 124 (62%). Notably, the spoken–written overlap is also visible for shorter trees, typically examined in related work, such as subtrees with up to three nodes, further underscoring the importance of considering modality-specific effects.

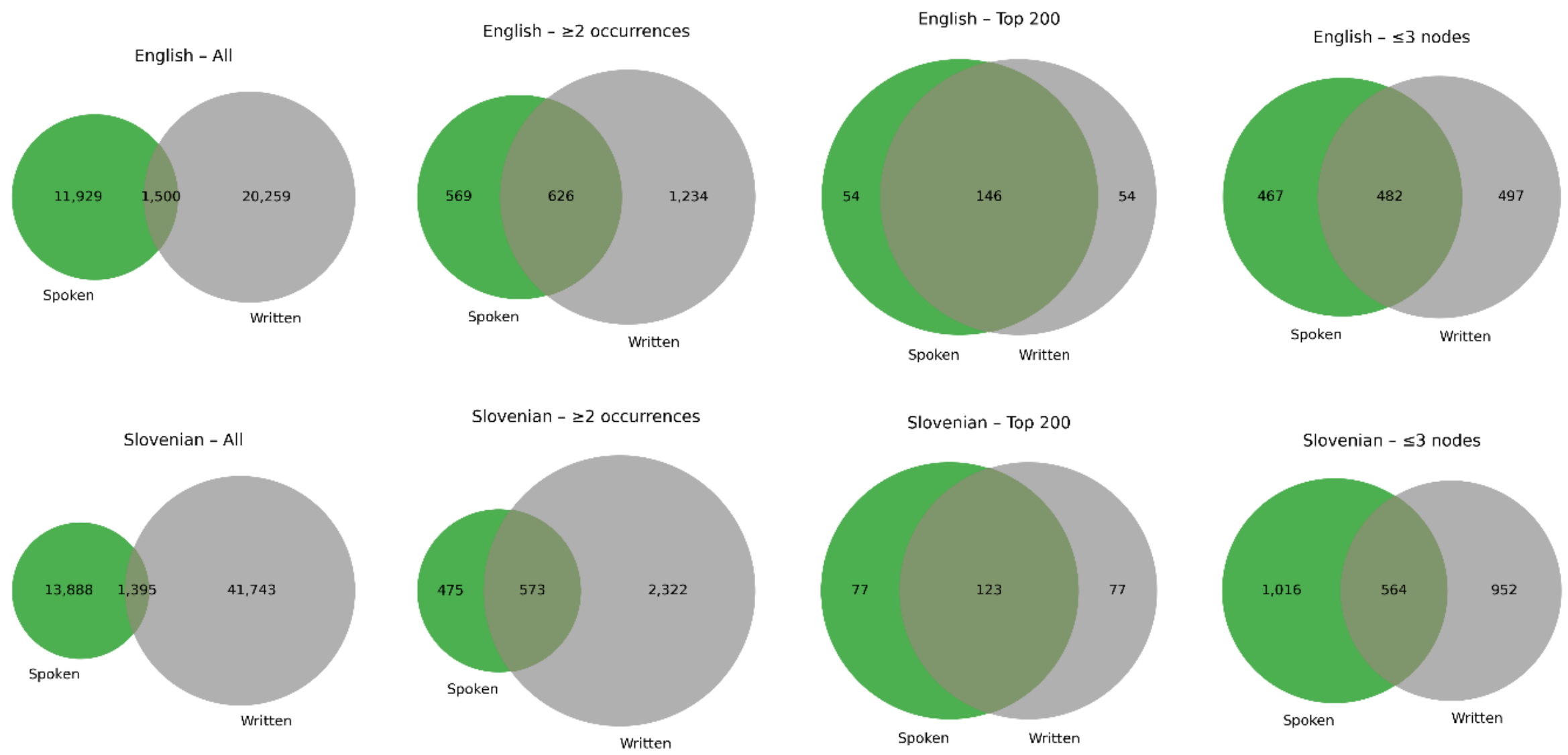

**Figure 7:** Overlap of syntactic structures between spoken and written corpora for English (top) and Slovenian (bottom) under four conditions: all structures, those attested at least twice, the 200 most frequent types, and structures with three nodes or less.

These findings confirm that spoken and written language draw on distinct subsets of syntactic resources, with only partial convergence. Crucially, this pattern is consistent across both languages—even when we narrow the comparison to core syntactic phenomena, excluding self-repairs, discourse markers, and other disfluencies (see Figure B4), and when part-of-speech distinctions are collapsed to focus on purely relational configurations (see Figure C3). This finding highlights the deep-rooted nature of syntactic variation between speech and writing, which persists across different levels of structural representation and even when features unique to spoken communication are removed.

## 4.3 Identification of key syntactic structures in speech

Building on previous findings about the distinct nature of syntactic inventories in speech and writing, we now demonstrate one possible application of the extracted structure lists: identifying which specific syntactic patterns are most characteristic of spoken language. To this end, we apply percentage difference (%DIFF), a commonly used effect-size metric in corpus linguistics that captures the proportional difference between the normalized frequencies of a linguistic phenomenon in two corpora. In practice, the highest %DIFF values often highlight structures that are entirely absent from writing, effectively surfacing constructions unique to speech. Tables 6 and 7 show the ten most distinctive syntactic structures for spoken English and Slovenian, respectively. These results are based on the structurally reduced version of the corpora excluding repairs and discourse markers (Section 2.4), which would otherwise dominate the top-ten lists (see Table D1 and D2 in Appendix D). All observed differences are statistically significant ($p < 0.001$ based on log-likelihood). We emphasize that this keyness-based analysis represents just one of many possible ways to explore the extracted inventories; alternative keyness metrics, filtering criteria, or analytical foci would surface different patterns and insights, as discussed in Section 5.

**Table 6:** Top-10 syntactic structures in spoken English, ranked by %DIFF keyness measure. Frequency values shown for spoken (GUM-S) and written (GUM-W) corpora are absolute (raw counts).

| | Tree (with example) | GUM-S | GUM-W | %DIFF |
|---|---|---|---|---|
| 1 | PRON <nsubj AUX <aux VERB >advmod ADV<br>***(You could come too)*** | 13 | 0 | 2.20E+21 |
| 2 | AUX >nsubj PRON<br>*(You're not gonna let me finish **are you**)* | 12 | 0 | 2.03E+21 |
| 3 | VERB >obj PRON >xcomp VERB<br>***(Let's see)*** | 9 | 0 | 1.52E+21 |
| 4 | SCONJ <mark PRON <nsubj AUX >advmod PART<br>*(**If you don't** I will)* | 8 | 0 | 1.35E+21 |
| 5 | PRON <nsubj AUX <cop ADV <advmod ADV <advmod ADJ<br>***(It's really really thick)*** | 7 | 0 | 1.18E+21 |
| 6 | ADP <case PRON <dep NOUN<br>*(I wanted to get these out **to you guys**)* | 6 | 0 | 1.01E+21 |
| 7 | DET <det ADJ <amod NOUN >nmod (ADP <case PRON)<br>*(And **the first part of it** is we have lecturethen we have lab)* | 6 | 0 | 1.01E+21 |
| 8 | `PRON <nsubj AUX <cop NOUN`<br>*(**That is Dad**)* | 6 | 0 | 1.01E+21 |
| 9 | `PRON <nsubj VERB >xcomp (PART <mark VERB >obj PRON)`<br>*(I don't have any this year **I forgot to plant it**)* | 6 | 0 | 1.01E+21 |
| 10 | `NOUN >acl PART`<br>(*high quality* ***how to*** *information*) | 6 | 0 | 1.01E+21 |

**Table 7:** Top-10 syntactic structures in spoken Slovenian, ranked by %DIFF keyness measure. Frequency values shown for spoken (SST) and written (SSJ) corpora are absolute (raw counts).

| | **Tree (with example)** | **SST** | **SSJ** | **%DIFF** |
|---|---|---|---|---|
| 1 | `(CCONJ >fixed CCONJ) <cc ADJ`<br>(*lahko so to slani* ***ali pa sladki,*** 'they can be salty or sweet') | 11 | 0 | 3.66E+21 |
| 2 | `(CCONJ >fixed CCONJ) <cc NOUN`<br>(*čevlje nogavice* ***ali pa kravato,*** **'**shoes socks or a tie') | 11 | 0 | 3.66E+21 |
| 3 | `DET <nsubj AUX <cop DET`<br>*(zraka ne vsebuje **to je tisto**,* 'it has no air that's it') | 10 | 0 | 3.33E+21 |
| 4 | `ADP <case VERB`<br>(*prijeten park* ***za sprehajati*** *je,* 'it's a nice park to walk in') | 9 | 0 | 3.00E+21 |
| *5* | `ADV >orphan CCONJ`<br>(***kam pa,*** 'where to') | 9 | 0 | 3.00E+21 |
| 6 | `ADV >vocative PROPN`<br>(***naprej [name:person],*** 'next [name]') | 7 | 0 | 2.33E+21 |
| 7 | `DET >amod ADJ`<br>(*ampak želim* ***nekaj drugega*** *poudariti,* 'I want to emphasize something else') | 7 | 0 | 2.33E+21 |
| 8 | `PRON >advmod PART`<br>(*pa imam* ***jaz tudi*** *svoje obveznosti,* 'I also have my responsibilities') | 7 | 0 | 2.33E+21 |
| 9 | `VERB >ccomp SCONJ`<br>(***upam da,*** 'I hope so') | 7 | 0 | 2.33E+21 |
| 10 | `VERB >parataxis VERB`<br>(***napiši daj,*** 'write go on') | 7 | 0 | 2.33E+21 |

Although limited to just ten entries, the top structures in spoken English (Table 6) already point to several characteristic patterns. Many involve short pronominal clauses (*you could come too, that is Dad, I forgot to plant it*), elliptical constructions (*if you don't*), and formulaic expressions (*let's see*), as well as interrogative patterns (*are you?*), which reflect the interactive, context-grounding, and economical dimensions of speech. The list also features constructions with colloquial multi-word expressions (*you guys, how-to*) and repetition-based intensification (*really really thick*). Notably, when part-of-speech distinctions are collapsed in English (Table C4 in Appendix C), similar patterns persist among the top-ranked structures, confirming that these speech-specific preferences reflect underlying syntactic configurations rather than being driven solely by differences in vocabulary or word category choices, such as the greater prominence of pronouns over nouns shown in Figure 6.

The top-ranking speech-specific structures in Slovenian (Table 7) show similar tendencies. These include structures with the colloquial multi-word conjunction *ali pa* 'or', formulaic expressions like *to je tisto* 'that's it', and several cases of predicate ellipsis in questions (*kam pa?* 'where to?'), replies (*upam da* 'I hope so'), and imperatives (standalone *naprej* 'next!'*;* coordinated *napiši, daj* 'write, go on!'), as hinted at by the dependent-less clausal heads. Additional patterns include atypical intensifier order in *jaz tudi* 'I also'—where *tudi jaz* 'also I' would be more expected in writing—and infinitival constructions with prepositions (*za sprehajati* 'to walk'), which are typical of informal Slovenian. When part-of-speech distinctions are collapsed (Table C5 in Appendix C), the top-ranked structures differ somewhat, revealing additional examples of atypical word order, such as post-verbal consecutive adverbials (*hoditi tolikokrat tja* 'go there so often'), emphatic pre-infinitive object placement (*se da tudi to zmagati* 'this can also be won'), and post-verbal subjects (*odštevajo se sekunde* 'seconds are counted'). Additionally, new types of formulaic expressions emerge, such as general extenders (*ali pa kaj drugega* 'or something else'*, in tako naprej* 'and so on')—alongside some annotation artifacts (e.g., *goeswith* for word segmentation errors, *parataxis* for certain appositive constructions) that stem from corpus-specific annotation decisions, highlighting the method's additional benefit as a tool for annotation consistency control**.**

Taken together, these findings show that the most distinctive speech-specific structures are largely pragmatically motivated, reflecting known dimensions of spoken language production such as interactivity, context-dependence, and economy of expression. While these properties have been well documented in previous work, the patterns identified in Tables 6 and 7 demonstrate that such constructions can be systematically and comprehensively extracted using a fully data-driven, language-independent approach. Regardless of the level of structural representation, the resulting inventories of structures opens new avenues for researchers to either extend the analysis to less frequent patterns or apply other statistical methods to identify speech-specific structures and explore additional dimensions of syntactic variation.

# 5 Discussion

This study develops a systematic method for bottom-up analysis of syntactic variation across corpora, using fully structured dependency trees as the unit of comparison. As such, it aligns with the growing trend of using syntactically parsed corpora for linguistic research, covering diverse research areas from typology (e.g. Gerdes et al. 2021; Levshina 2022; Levshina et al. 2023) to applied linguistics (e.g. Díez-Bedmar and Pérez-Paredes 2020; Jiang et al. 2019; Kyle and Crossley 2017; Lu 2010). However, unlike prevailing treebank-based approaches to investigating syntactic variation, which often focus on analyzing the distribution of specific syntactic types, functions, or patterns (as discussed in the Introduction), our fully inductive, 'treebank-driven' method captures hierarchical syntactic structures of any size or type, without imposing prior assumptions about what counts as linguistically relevant. Its foundation in dependency grammar makes it inherently language-independent and scalable, accommodating diverse word order configurations and enabling systematic, data-driven investigations of grammatical usage beyond the limitations of traditional corpus queries.

Applying this method to spoken and written corpora of English and Slovenian, we observed consistent and meaningful contrasts across three dimensions. First, spoken corpora showed lower structural diversity compared to written corpora—confirming that spoken language tends to rely more on formulaic and structurally repetitive patterns (Biber et al. 1999, Erman and Warren 2000). Second, we discovered that the syntactic inventories of speech and writing are not only different in their distribution but also surprisingly distinct, with only partial overlap—even when focusing on frequent or repeated structures. Third, our keyness analysis identified specific syntactic structures disproportionately frequent in speech, including short elliptical clauses, formulaic expressions, colloquial constructions, and other configurations closely tied to the specific communicative demands of real-time interaction.

Indeed, while many of our findings on the distinct nature of spoken language echo earlier observations in seminal work by Biber (1988) and have been reinforced by numerous studies since, this study makes two crucial contributions that advance the field. First, it systematically defines and quantifies a complete, finite inventory of fully structured syntactic patterns, enabling fine-grained, bottom-up comparisons across modalities and uncovering patterns of syntactic variation that are not easily accessible through more traditional, top-down approaches. Second, it highlights the considerable potential of this method to uncover both universal and language-specific grammatical patterns in speech – especially given the growing availability of spoken language corpora annotated using the Universal Dependencies scheme (Dobrovoljc 2022). By demonstrating the method's applicability across typologically distinct languages like English and Slovenian, this study reveals a surprising consistency in spoken-language traits while also surfacing language-specific differences, opening new avenues for systematic cross-linguistic and cross-modal research.

At the same time, this study represents only an initial exploration of what the method makes possible. With over 300 treebanks now available in the Universal Dependencies framework, covering more than 180 languages and a wide range of genres (Zeman et al. 2025), and with the increasing availability of high-accuracy parsers such as Stanza (Qi et al. 2020) and Trankit (Nguyen et al. 2021), this approach is directly applicable to a vast and continually expanding range of corpora. Moreover, its configurability opens the door to more focused analyses: by adjusting tree granularity—such as integrating morphosyntactic features for finer distinctions, removing part-of-speech or dependency relation information to focus on abstract structural patterns, or adding lexical information to explore lexico-grammatical configurations (Paquot et al. 2020)—researchers can tailor the approach to a broad array of linguistic questions. Importantly, while this study used complete dependency trees and subtrees as units of analysis, these could also be redefined to include all possible head-dependent configurations (paths) allowing for even more granular or targeted investigations of syntactic patterns.

Similarly, the scope of analysis can be flexibly adapted at multiple stages. At the extraction stage, the method can be narrowed from broad syntactic comparisons, as presented in this paper, to more detailed—and, arguably, more compelling—explorations of specific phrasal or clausal types of trees, enabling direct comparisons with established corpus-based grammar descriptions (Biber et al. 1999; Carter and McCarthy 2006; Hunston and Francis 2000; Sinclair and Maureen 2006), or their extension to languages beyond English. At the analysis stage, researchers can apply various post-extraction procedures: the keyness analysis in Section 4.3, for example, represents just one exploratory application of the extracted inventories. Alternative keyness metrics, filtering criteria, or analytical foci could surface different patterns depending on research questions—for instance, researchers interested specifically in spoken language might further distinguish pronoun- and noun-headed trees, introduce specific morphological features as tree-distinguishing variables, or exclude formulaic relations like *fixed* to better isolate productive syntactic variation.

More broadly, by introducing dependency (sub)trees as quantifiable units of analysis, this approach extends the reach of corpus-driven, inductive statistical techniques—previously focused mainly on lexical and other surface-level phenomena—to the domain of syntactic structures as well. These include alternative measures of keyness (Bondi and Scott 2010;

Gabrielatos 2018), word association (Evert 2009), clustering (Moisl 2015), collocation networks (Brezina et al. 2015), and multidimensional analysis (Biber 2017)—to mention just a few data-driven distributional methods that identify salient patterns directly from corpus data.

In sum, the flexibility of the proposed method, combined with its scalability grounded in cross-linguistically consistent UD annotations, paves the way for fully data-driven, cross-linguistic, and richly nuanced investigations of syntactic variation—laying the foundation for new empirical discoveries and, ultimately, new theories of syntax in use.

# 6 Conclusion

This paper introduced a systematic, bottom-up method for identifying and comparing syntactic structures in spoken and written language, based on the exhaustive extraction of delexicalized dependency trees from parsed corpora. Applied to English and Slovenian, the approach revealed consistent cross-modal contrasts in syntactic inventory size, diversity, and overlap—providing rare cross-linguistic evidence of modality-specific structural preferences in speech. All corpora, configuration files, and extracted outputs are openly available, ensuring that the method is transparent, reproducible, and ready to be extended by others, with genre-controlled comparisons representing a particularly promising avenue for future research.

# A. Overview of Universal Dependencies POS tags and dependency relations

**Table A1:** List of universal part-of-speech tags in the UD annotation scheme.

| Tag | Meaning | Tag | Meaning |
|---|---|---|---|
| ADJ | Adjective | PART | Particle |
| ADP | Adposition (preposition/postposition) | PRON | Pronoun |
| ADV | Adverb | PROPN | Proper noun |
| AUX | Auxiliary verb | PUNCT | Punctuation |
| CCONJ | Coordinating conjunction | SCONJ | Subordinating conjunction |
| DET | Determiner | SYM | Symbol |
| INTJ | Interjection | VERB | Main verb |
| NOUN | Common noun | X | Other/unknown |
| NUM | Numeral | | |

**Table A2:** List of universal dependency relations in the UD annotation scheme.

| Label | Meaning | Label | Meaning |
|---|---|---|---|
| acl | Clausal modifier of noun | fixed | Fixed multiword expression |
| advcl | Adverbial clause modifier | flat | Flat multiword expression |
| advmod | Adverbial modifier | goeswith | Partial word |
| amod | Adjectival modifier | iobj | Indirect object |
| appos | Appositional modifier | list | List element |
| aux | Auxiliary | mark | Marker of subordinate clause |
| case | Case marker (preposition) | nmod | Nominal modifier |
| cc | Coordinating conjunction | nsubj | Nominal subject |
| ccomp | Clausal complement | nummod | Numeric modifier |
| clf | Classifier | obj | Object |
| compound | Compound noun modifier | obl | Oblique nominal |
| conj | Conjunct in coordination | orphan | Orphan in elliptical structures |
| cop | Copula | parataxis | Parataxis |
| csubj | Clausal subject | punct | Punctuation |
| dep | Unspecified dependency | reparandum | Overridden disfluency |
| det | Determiner | root | Root of the sentence |
| discourse | Discourse element | vocative | Vocative expression |
| dislocated | Dislocated argument | xcomp | Open clausal complement |
| expl | Expletive | | |

## B. Results for disfluency-free trees

**Table B1:** Overview of extracted syntactic structure inventories in English and Slovenian spoken and written corpora (disfluency-free versions): tokens, types, and hapax legomena.

| | **English** | | **Slovenian** | |
|---|---|---|---|---|
| | **Spoken** | **Written** | **Spoken** | **Written** |
| **Structure tokens** | 67,031 | 113,199 | 68,281 | 227,421 |
| **Structure types** | 12,856 | 21,717 | 14,198 | 43,114 |
| **Hapax structures** | 11,680 | 19,857 | 13,215 | 40,220 |

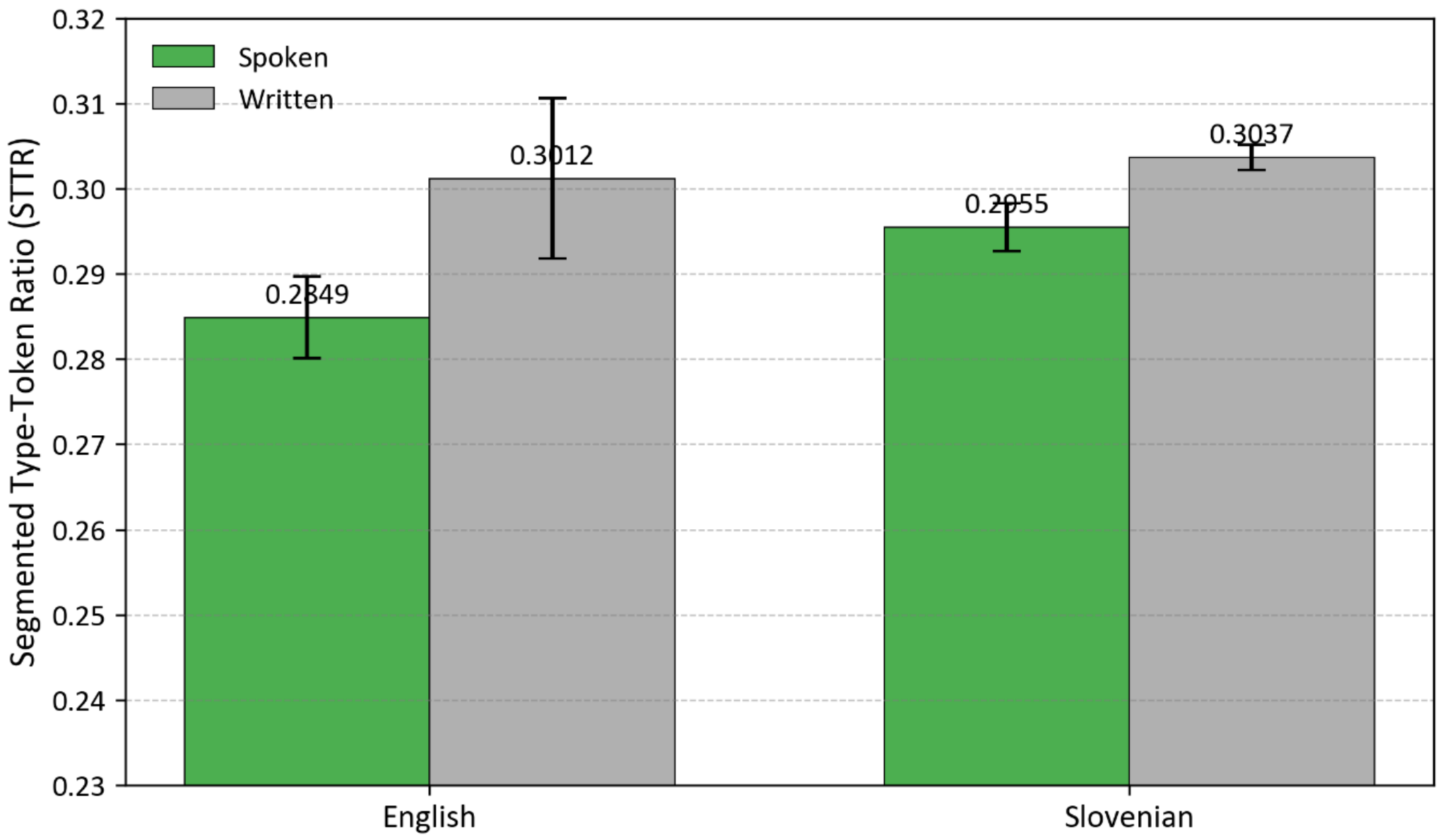

**Figure B2:** Segmented Type-Token Ratio (STTR) in spoken and written treebanks for Slovenian and English (disfluency-free versions). Bars show average TTR values across 1,000-token segments; error bars indicate 95% confidence intervals. (Note: The inflated confidence interval for written English results from a 19-token outlier segment with abnormally high STTR, which remains included for transparency.)

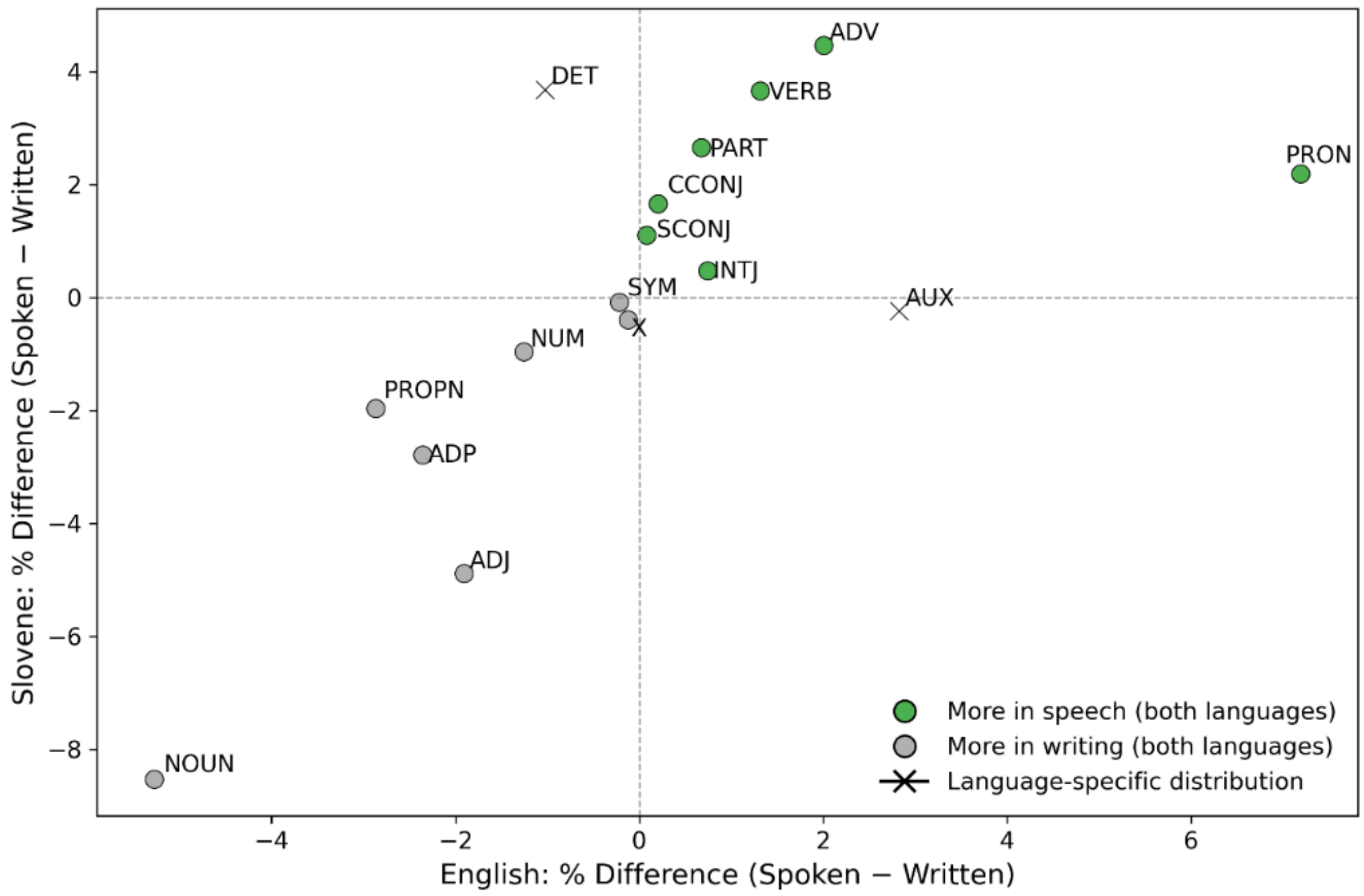

**Figure B3:** Cross-linguistic comparison of structure distributions in spoken and written corpora (disfluency-free versions) according to structural type (head POS). Positive values indicate higher frequency in speech; negative values indicate higher frequency in writing.

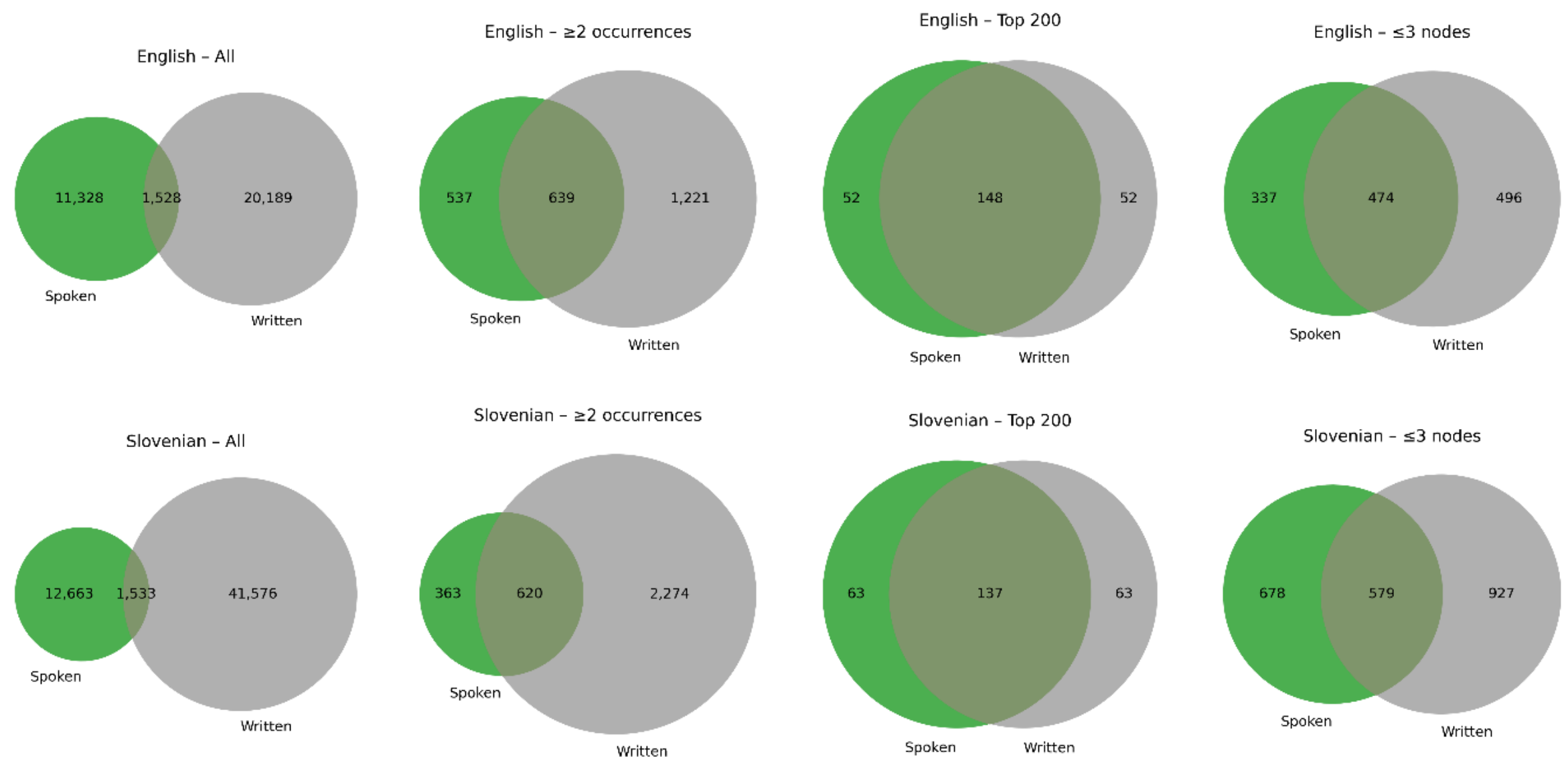

**Figure B4:** Overlap of syntactic structures in the disfluency-free versions of the spoken and written corpora for English (above) and Slovenian (below) under four conditions: all structures, structures attested at least twice, the 200 most frequent types, and structures with three nodes or less.

## C. Results for dependency-only trees

**Table C1:** Overview of extracted syntactic structure inventories in English and Slovenian spoken and written corpora, using dependency-only tree representations (relation labels without node part-of-speech information): tokens, types, and hapax legomena.

| | English | | Slovenian | |
|---|---|---|---|---|
| | **Spoken** | **Written** | **Spoken** | **Written** |
| **Structure tokens** | 67,031 | 113,199 | 68,281 | 227,421 |
| **Structure types** | 12,856 | 21,717 | 14,198 | 43,114 |
| **Hapax structures** | 11,680 | 19,857 | 13,215 | 40,220 |

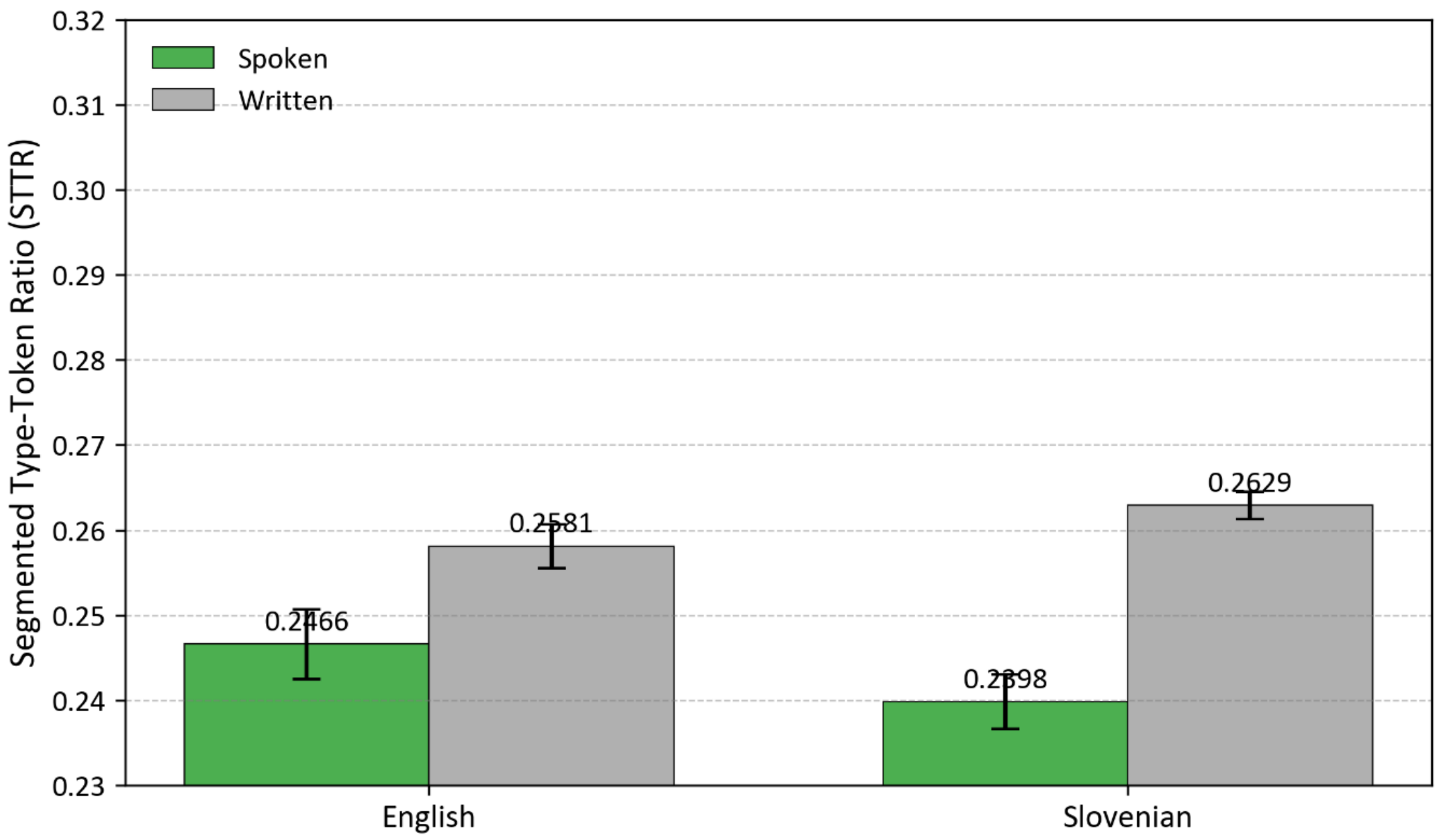

**Figure C2:** Segmented Type-Token Ratio (STTR) in spoken and written treebanks for Slovenian and English (dependency-only tree representation). Bars show average TTR values across 1,000-token segments; error bars indicate 95% confidence intervals.

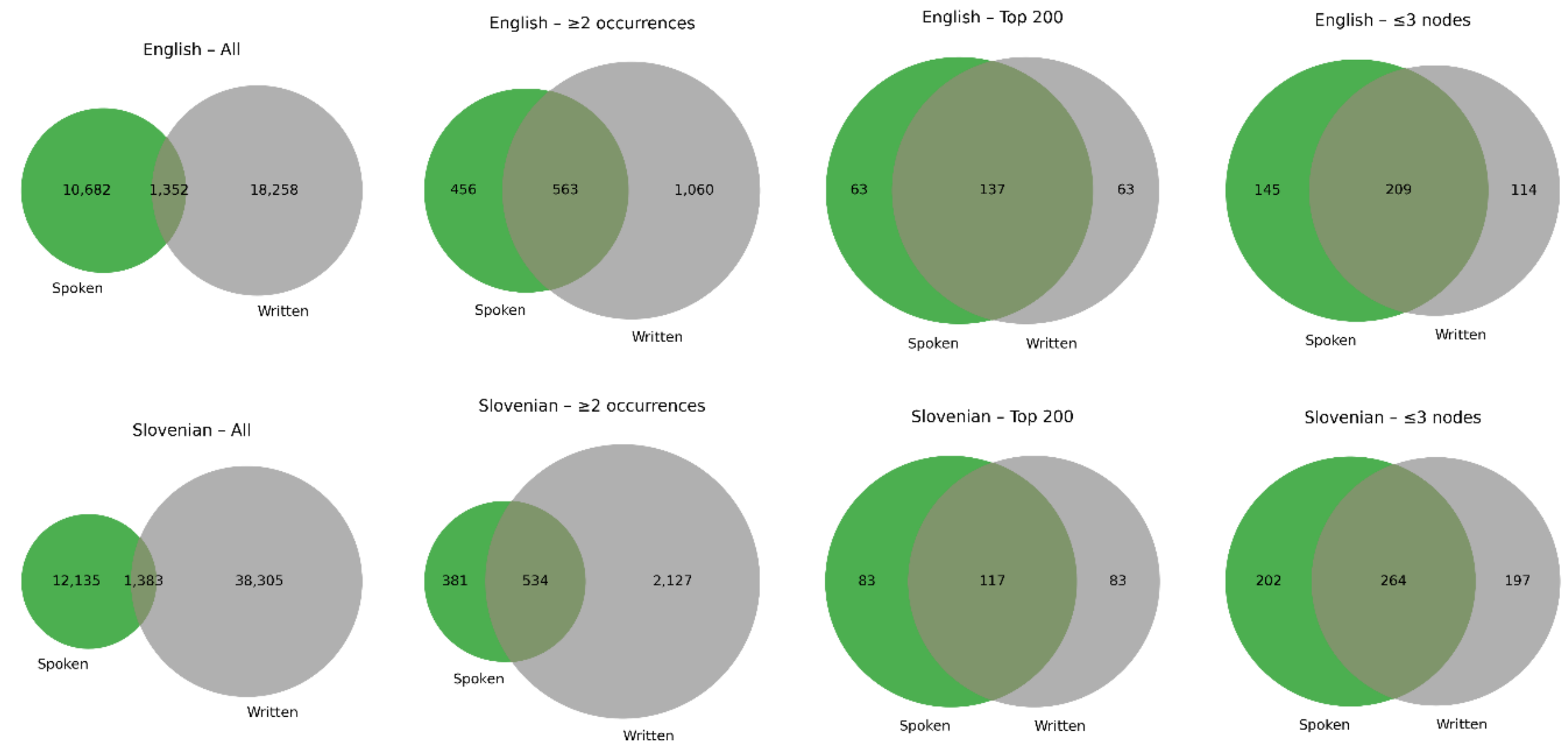

**Figure C3:** Overlap of syntactic structures in English (top) and Slovenian (bottom) spoken and written corpora, based on dependency-only tree representations, under four conditions: all structures, structures attested at least twice, the 200 most frequent types, and structures with three nodes or less.

**Table C4:** Top 10 speech-specific structures in English (dependency-only representation), ranked by %DIFF. For comparability with Section 4.3, the list excludes structures featuring relations pertaining to disfluencies (e.g., *discourse* and *reparandum*).

| **Tree** | **GUM-s** | **GUM-w** | **%DIFF** |
|---|---|---|---|
| `_ <nsubj _ <aux _ >advmod _`<br>*(I think **I'll stop there**)* | 11 | 0 | 1,79E+21 |
| `_ >obj _ >compound _`<br>***(Pull em out)*** | 7 | 0 | 1,14E+21 |
| `_ <expl _ >nsubj (_ <det _)`<br>***(There was a connection)*** | 6 | 0 | 9,77E+20 |
| `_ <mark _ <nsubj _ <aux _ >xcomp (_ <mark _ >obj _)`<br>***(Because we'd have to undo it)*** | 6 | 0 | 9,77E+20 |
| `_ <nsubj _ <cop _ >acl (_ <nsubj _)`<br>***(That's what I assumed)*** | 6 | 0 | 9,77E+20 |
| `_ >cop _ >nsubj _`<br>***(How are you)*** | 6 | 0 | 9,77E+20 |
| `_ <advmod _ <aux _ <nsubj _ >obj _`<br>(***How did he explain it)*** | 5 | 0 | 8,14E+20 |
| `_ <aux _ <nsubj _ >obj (_ <det _)`<br>***(Do you need a partner)*** | 5 | 0 | 8,14E+20 |
| `_ <det _ >acl (_ <obj _ <nsubj _)` | 5 | 0 | 8,14E+20 |
| *(So Erasmus laid the the laid* ***the egg which Luther hatched)*** | | | |
| _ <advmod _ <aux _ <nsubj _<br>***(Where do you advertise)*** | 4 | 0 | 6,51E+20 |

**Table C5:** Top 10 speech-specific structures in Slovenian (dependency-only representation), ranked by %DIFF. For comparability with Section 4.3, the list excludes structures featuring relations pertaining to disfluencies (e.g., *discourse* and *reparandum*).

| **Tree** | **SST** | **SSJ** | **%DIFF** |
|---|---|---|---|
| _ >parataxis _<br>(***ni ni***, 'it-isn't, it-isn't') | 10 | 0 | 2,98E+21 |
| _ >advmod _ >advmod _<br>*(Nisem mogel* ***hoditi tolikokrat tja,*** 'I couldn't go there so-often') | 8 | 0 | 2,39E+21 |
| (_ >fixed _) <cc _ >nmod _<br>(***Ali pa kaj drugega***, 'or something else') | 6 | 0 | 1,79E+21 |
| _ >goeswith _<br>(*a* ***ni smo*** *šli pol na ono stran dol*, 'we did n't go to the other side then') | 6 | 0 | 1,79E+21 |
| (_ <advmod _) <obj _<br>(*eee se da* ***tudi to zmagati***, 'eer *this* can *also* *be won'*) | 5 | 0 | 1,49E+21 |
| _ >expl _ >nsubj _<br>(***odštevajo se sekunde,*** 'seconds are counted') | 5 | 0 | 1,49E+21 |
| _ <mark _ <aux _ <iobj _<br>(*pa si kdaj študiral* ***da bi koga učil***, 'did you ever consider to teach someone') | 4 | 0 | 1,19E+21 |
| _ <parataxis _ <case _<br>(*vsaj v primerjavi z nekaterimi sosednjimi državami* ***recimo z Italijo***, 'at least in comparison to some neighbourring countries say to Italy') | 4 | 0 | 1,19E+21 |
| ***_ >conj (_ >fixed _ >fixed _)***<br>(*prepleteni* ***in tako naprej***, 'interwined and so on') | 4 | 0 | 1,19E+21 |
| *_ >conj _ >conj ((_ >fixed _) <cc _)*<br>(zadostno količino **moke, jajc in pa mleka**, 'sufficient quantity of flour, eggs and milk') | 4 | 0 | 1,19E+21 |

## D. Key structures in main corpora (punctuation-free)

**Table D1:** Top 10 speech-specific syntactic structures in English (punctuation-free corpus), ranked by %DIFF keyness measure.

| Tree | GUM-s | GUM-w | %DIFF |
|---|---|---|---|
| DET <reparandum DET | 19 | 0 | 3.09E+21 |
| INTJ >discourse INTJ | 15 | 0 | 2.44E+21 |
| AUX >nsubj PRON | 13 | 0 | 2.12E+21 |
| CCONJ <reparandum CCONJ | 13 | 0 | 2.12E+21 |
| SCONJ <mark PRON | 12 | 0 | 1.95E+21 |
| PRON <nsubj AUX <aux VERB >advmod ADV | 11 | 0 | 1.79E+21 |
| INTJ <discourse PRON <nsubj AUX <cop ADJ | 11 | 0 | 1.79E+21 |
| VERB >obj PRON >xcomp VERB | 9 | 0 | 1.47E+21 |
| INTJ <reparandum INTJ | 9 | 0 | 1.47E+21 |
| INTJ <reparandum PRON | 9 | 0 | 1.47E+21 |

**Table D2:** Top 10 speech-specific syntactic structures in Slovenian (punctuation-free corpus), ranked by %DIFF keyness measure.

| Tree | SST | SSJ | %DIFF |
|---|---|---|---|
| PART >discourse PART | 54 | 0 | 1.61E+22 |
| DET <reparandum DET | 53 | 0 | 1.58E+22 |
| ADP <reparandum ADP | 52 | 0 | 1.55E+22 |
| CCONJ <reparandum CCONJ | 49 | 0 | 1.46E+22 |
| SCONJ <reparandum SCONJ | 47 | 0 | 1.40E+22 |
| X <reparandum ADV | 36 | 0 | 1.07E+22 |
| X <reparandum ADJ | 30 | 0 | 8.94E+21 |
| ADV <reparandum ADV | 27 | 0 | 8.05E+21 |
| SCONJ <mark AUX | 22 | 0 | 6.56E+21 |
| INTJ >discourse INTJ | 21 | 0 | 6.26E+21 |

## Acknowledgment

This work was financially supported by the Slovenian Research and Innovation Agency through the research project SPOT: A Treebank-Driven Approach to the Study of Spoken Slovenian (Z6-4617, https://spot.ff.uni-lj.si/), and the research program Language Resources and Technologies for Slovene (P6-0411). Constructive feedback on earlier versions of this work was provided by participants of Språkvetenskapligt forum at the University of Gothenburg and by members of the COST Action UniDive CA21167, whose input is gratefully acknowledged. Thanks are also due to the anonymous reviewers for their insightful and constructive comments, which greatly helped improve the clarity and precision of this paper. Generative AI tools were employed to support language editing. The author retains full responsibility for the content, analysis, and interpretation presented in this work.